\pdfoutput=1

\documentclass[11pt]{article}

\usepackage{EMNLP2022}

\usepackage{times}
\usepackage{latexsym}

\usepackage[T1]{fontenc}

\usepackage[utf8]{inputenc}

\usepackage{microtype}

\usepackage{inconsolata}
\usepackage{booktabs}
\usepackage{graphicx}

\usepackage{amsmath, amsfonts}
\usepackage{algorithm}
\usepackage{algpseudocode}
\usepackage{array}

\title{Training Large-Vocabulary Neural Language Models by \\
	   Private Federated Learning for Resource-Constrained Devices}

\author{Mingbin Xu, Congzheng Song \\ 
	{\bf Ye Tian, Neha Agrawal, Filip Granqvist, Rogier van Dalen, Xiao Zhang} \\ 
	{\bf Arturo Argueta, Shiyi Han, Yaqiao Deng, Leo Liu, Anmol Walia, Alex Jin} \\
	Apple Inc. \\
	\texttt{\{mingbin.xu, csong4, ytian2, neha\_agrawal, fgranqvist, rogier\_vandalen,} \\
	\texttt{xiaoz, arturo\_argueta, shan26, yaqiao\_deng, lliu9, awalia, alex\_jin\}@apple.com}
}

\begin{document}
\maketitle
\begin{abstract}
Federated Learning (FL) is a technique to train models using data distributed across devices.
Differential Privacy (DP) provides a formal privacy guarantee for sensitive data. 
Our goal is to train a large neural network language model (NNLM) on compute-constrained devices while preserving privacy using FL and DP. 
However, the DP-noise introduced to the model increases as the model size grows, which often prevents convergence.
We propose Partial Embedding Updates (PEU), a novel technique to decrease noise by decreasing payload size. 
Furthermore, we adopt Low Rank Adaptation (LoRA) and Noise Contrastive Estimation (NCE) to reduce the memory demands of large models on compute-constrained devices. 
This combination of techniques makes it possible to train large-vocabulary language models while preserving accuracy and privacy.


\end{abstract}

\section{Introduction}

Language Models (LMs) play an important role in many applications, including Automatic Speech Recognition (ASR), Machine Translation (MT), Information Retrieval (IR), and Natural Language Understanding (NLU). 
The transformer-based \cite{DBLP:conf/nips/VaswaniSPUJGKP17} neural network language model (NNLM) launched an era of large-scale LM pre-training and task-specific parameter fine-tuning \cite{DBLP:conf/naacl/DevlinCLT19, radford2018improving, radford2019language, DBLP:conf/nips/BrownMRSKDNSSAA20, DBLP:journals/corr/abs-1909-08053}. 

Data stored on user's device is private, and cannot be transmitted to the server.
\citet{DBLP:conf/aistats/McMahanMRHA17} present an federated learning (FL) method that trains a global model from a federation of edge devices that transmit gradients - instead of raw data - to the server.
However, mobile devices are constrained in terms of memory and bandwidth. 
Many applications require a prohibitively large number of classes to model labels in the tail of the distribution. 
Those tail cases are of greater interest from the user experience perspective \cite{DBLP:conf/interspeech/SaebiPMG21}.

Lack of access to the raw data does not necessarily guarantee privacy. 
Model gradients may reveal some information about user data~\cite{DBLP:conf/sp/MelisSCS19,DBLP:conf/nips/ZhuLH19}.
Neural networks tend to memorize training data ~\cite{DBLP:conf/uss/Carlini0EKS19,DBLP:conf/uss/CarliniTWJHLRBS21}. 
While differential privacy (DP) \cite{DBLP:conf/tcc/DworkMNS06} meets the privacy demand of SGD-based algorithm \cite{DBLP:conf/ccs/AbadiCGMMT016}, the DP-perturbation can quickly overwhelm the gradient signal as the model capacity grows, which creates a dilemma between accuracy and privacy \cite{bassily2014private}.

In this work, differential privacy is used on top of federated learning to learn task-specific LMs while guaranteeing privacy. 
We address resource limitations in federated learning with partial embedding updates, low-rank adaptation, and noise contrastive estimation, which co-jointly overcome the decreasing gradient utilization due to the increasing model size in the differentially private context.
We stabilize the training process and speed up convergence by exponentially decaying the moving average of the models.

\section{Preliminary}

\subsection{Federated Learning}
\label{sec:fl}

Federated Learning (FL)~\cite{DBLP:conf/aistats/McMahanMRHA17} was introduced to train a user-independent model with data from multiple users while ensuring personal data never leaves the device.
At iteration $t$ of FL, the server first samples a subset of devices $\mathcal{C}$ from the population.
Each sampled device $i\in\mathcal{C}$ then downloads the shared model parameters $\theta_t$ from the server and locally trains the model on its own data to produce a local model $\theta_i$.
The model difference $\Delta_{t,i} = \theta_{i} - \theta_t$ is sent back to the server. 
The server aggregates the model differences as a ``pseudo-gradient'' $\Delta_t=\frac{1}{|\mathcal{C}|}\sum_{i\in\mathcal{C}}\Delta_{t,i}$ and uses the peseudo-gradient to update $\theta_t$ with any standard optimizer~\cite{DBLP:conf/iclr/ReddiCZGRKKM21}.

\subsection{Differential Privacy}

Differential privacy (DP) defines a privacy guarantee for SGD-based learning algorithms.
DP is formally defined as follows:

\newtheorem{definition}{Definition}

\begin{definition}{Differential Privacy \cite{DBLP:conf/tcc/DworkMNS06}}
A randomized algorithm $\mathcal{A}:\mathcal{D}\mapsto\mathcal{R}$ is ($\epsilon,\delta$)-differentially private, if for any pair of neighboring training populations $\mathcal{D}$ and $\mathcal{D^\prime}$ and for any subset of outputs $\mathcal{S}\subseteq\mathcal{R}$, it holds that
\begin{equation}
	\mathrm{Pr}[\mathcal{A}(\mathcal{D})\in\mathcal{S}] \leq e^\epsilon\cdot \mathrm{Pr}[\mathcal{A}(\mathcal{D^\prime})\in\mathcal{S}] + \delta.
\label{eq:dp}
\end{equation}
\end{definition}
Training population $\mathcal{D^\prime}$ is considered the neighbor of $\mathcal{D}$ if $\mathcal{D^\prime}$ can be obtained by adding or removing one user from $\mathcal{D}$, and vice versa.
The choice of $(\epsilon, \delta)$ is a trade-off between accuracy and privacy. 

\citet{DBLP:conf/iclr/McMahanRT018} proposed building a FL framework with DP, which moves the trust barrier to the user devices. 
The interpretation of Equation \ref{eq:dp} in the FL setup is that the participation of an individual user and its contribution cannot be detected or reverse-engineered by any means.
We call such systems Private Federated Learning (PFL) hereafter.
The Gaussian mechanism \cite{DBLP:journals/fttcs/DworkR14} is widely adopted in PFL systems due to its simplicity and effectiveness.
On top of FL (\ref{sec:fl}), for each iteration:
\begin{enumerate}
	\item Each sampled client's model difference $\Delta_{t,i}$ is clipped to a zero-centered $L_2$ ball with radius $S$ so as to bound the sensitivity of $\Delta_{t,i}$. 
	\item A Gaussian noise with covariance matrix $\sigma^2\mathcal{\textbf{I}}$ is added to aggregation of $\sum_{i\in\mathcal{C}}\Delta_{t,i}$.
\end{enumerate}
The noise standard deviation is calibrated by the moment accountant~\cite{DBLP:conf/ccs/AbadiCGMMT016,DBLP:conf/csfw/Mironov17,DBLP:journals/corr/abs-1908-10530} with fixed sampling rate $q$ (fraction of clients sampled in each iteration), number of training iterations $T$, and privacy budgets $(\epsilon,\delta)$.

\section{Methods}



\subsection{Partial Embedding Updates}


Word frequency in natural language is highly skewed and long-tailed, and signal-to-DP-noise ratio (SNR) for rare words is thus insignificant.
We propose partial embedding updates (PEU), a sampling-based payload for the word embeddings.
At a high level, the server samples $m$ words from the vocabulary. 
Each device trains the full model but only sends back the updates for the sampled words. 
DP noise is added to the update of these $m$ words. 
The update of unsampled words will be zero with no noise added.
Let
\begin{itemize}
	\item $R(w)$ be the probability of word $w$ being sampled in each central iteration, 
	\item $Z \in \{0,1\}$ be a Bernoulli random variable where $P(Z = 1) = R(w)$, and
	\item $E_w$ and $E_w^0$ be the word vector of $w$ \textbf{after} and \textbf{before} local update respectively,
\end{itemize}
which allows us to rewrite the gradient of $w$ as
\begin{equation}
\begin{aligned}
	E_w - E_w^0 ={}& R(w) \frac{1}{R(w)} (E_w - E_w^0) \\
				={}& P(Z = 1) \cdot \frac{1}{R(w)} (E_w - E_w^0) \\
				 {}& + P(Z = 0) \cdot \textbf{0} \\
				={}& \mathbb{E}_{z\sim P}[\frac{z}{R(w)} (E_w-E_w^0)]
\end{aligned}
\label{eq:peu_exact}
\end{equation}
with $R(w) = P(w \in \mathcal{W})$ where $\mathcal{W}$ is a set of $m$ words sampled with some distribution $Q$ without replacement.
$R(w)$ does not seem to have a closed-form distribution if we choose a unigram distribution for $Q$.
To make the computation tractable, in the actual implementation shown in Algorithm \ref{alg:peu}, the expectation in Equation \ref{eq:peu_exact} is replaced by its Monte Carlo approximation and $Q$ is used to approximate $R$.

\begin{algorithm}[t]
  \caption{Partial Embedding Updates on Client}\label{alg:peu}
  \begin{algorithmic}[1]
    \Procedure{PEU}{$model, \mathcal{W}$}
      \State $\Delta_E, \Delta_{other} \gets SGD(model)$
      \State $\Delta_E^{\prime} \gets newArray()$
      \For{$w \in sampledWords$}
      	\State $append(\frac{1}{Q(w)}\Delta_{E_w}, \Delta_E^{\prime})$
      \EndFor
      \State $\Delta_{model} \gets concat(\Delta_E^{\prime}, \Delta_{other})$
      \State Send payload $\Delta_{model}$ to the server
    \EndProcedure
  \end{algorithmic}
\end{algorithm}

The dimension reduction of PEU benefits the PFL system in two important ways.
First, it decouples the model complexity requirement and the network requirement.
The number $m$ of sampled words can be determined by the payload budget.
Second, it increases the overall SNR because DP noise grows with the size of the model and tail words are less affected by DP noise.

\subsection{Low-Rank Adaptation}

\citet{DBLP:journals/corr/abs-2106-09685} developed LoRA, a generalized reparameterization approach to mitigate the immense memory requirement of fine-tuning large neural networks.
Since the major goal of LoRA is shrinking the trainable dimension, it benefits PFL in a similar fashion.
Prior to training, each dense matrix $W \in \mathbb{R}^{m \times n}$ is reformulated as a summation.
\begin{equation}
	W^{\prime} = W + L \cdot R
\label{eq:lora}
\end{equation}
where $L \in \mathbb{R}^{m \times r}$ is zero-initialized and $R \in \mathbb{R}^{r \times n}$ is randomly initialized. 
Typically $r$ is one or two orders of magnitudes smaller than $m$ and $n$.
$L$ and $R$ are trainable, and $W$ is frozen.

\subsection{Exponentially-decayed Moving Average}
\label{sec:ema}

Exponentially-decayed moving average (EMA) of trained model parameters often boosts the model performance in practice~\cite{DBLP:journals/jmlr/Martens20}. 
To apply EMA in private federated learning, the server maintains an additional set of model parameters $\phi$ and updates $\phi$ each central iteration after the model parameters $\theta$ are updated. 
Specifically, at central iteration $t$, $\phi$ is updated as 
\begin{equation}
\phi_{t+1} \gets\gamma\phi_t + (1-\gamma)\theta_{t+1}
\end{equation}
where $\gamma$ is the decaying rate between 0 and 1. The averaged parameters $\phi_t$ can potentially generalize better than model parameters $\theta_t$ at any single iteration $t$.
Analogous to momentum SGD, EMA of the model parameters stabilizes the DP-noise during training and leads to quicker and sometimes better convergence.

\subsection{Noise Contrastive Estimation}

\begin{table}
\centering
\begin{tabular}{p{0.15\textwidth}|p{0.12\textwidth}|p{0.12\textwidth}}
\toprule
\textbf{} & \textbf{SoftMax} & \textbf{NCE} \\
\midrule
embed param & 98 MB & 98 MB  \\
gradient &  98 MB & $<$ 1 MB\\
outputs & 200 MB & $<$ 5 MB \\
\midrule
\textbf{total} & $\approx$ 300 MB \footnote{We assume SoftMax is implemented as in-place as possible and reuses output buffer for backpropagation. The gap is much larger in practice.} & $\approx$ 100 MB \\
\bottomrule
\end{tabular}
\caption{Word embedding memory requirement between SoftMax and NCE, assuming a $10k \times 256$ word embedding and a $32 utterances/batch \times 16 words/utterance$ minibatch with vanilla SGD.}
\label{tab:nce-mem}
\end{table}

Noise contrastive estimation (NCE) \cite{DBLP:journals/jmlr/GutmannH10} is a sampling-based self-normalized approach that approximates SoftMax over large vocabulary efficiently \cite{DBLP:conf/icml/MnihT12}.
If a word $w$ is sampled from the corpus with auxiliary label $D = 1$, and $k$ other words are sampled from unigram $p_{uni}$ with auxiliary label $D = 0$.
The probability of $D$ is a mixture of $p_{\theta}(w, c)$ and $p_{uni}(w)$, where $p_{\theta}(w,c)$ represents the probability of the word $w$ given the context $c$, modeled by the LM parameter $\theta$:
\begin{equation}
\begin{aligned}
	p(D=0|c,w) &= \frac{k \times p_{uni}(w)}{p_{\theta}(w, c) + k \times p_{uni}(w)}\\
	p(D=1|c,w) &= \frac{p_{\theta}(w, c)}{p_{\theta}(w, c) + k \times p_{uni}(w)}
\end{aligned}
\label{eq:nce-prob}
\end{equation}
The LM distribution is learned by maximizing the log-likelihood of $D$:
\begin{equation}
\begin{aligned}
	Loss_{NCE_k} \approx {}& \displaystyle\sum_{(w,c)\in corpus}(log\ p(D=0|c,w) \\
					  {}& + \displaystyle\sum^k log\ p(D=0|c,\overline{w}))
\end{aligned}
\label{eq:nce-loss}
\end{equation}
$p_{\theta}(w, c)$ is the dot product of penultimate and $w$'s embedding without normalization.
While the initial purpose of NCE is to reduce computation, its memory saving is more appealing to PFL. 
The memory requirement for gradient and intermediate output are controllable by batch size and sampled words, independent of the embedding matrix size.

\begin{table}
\centering
\begin{tabular}{ll}
\toprule
$(\epsilon, \delta)$ & $(2, 10^{-6})$ \\
sampling rate & $2 \times 10^{-3}$ \\
${L_2}$ clipping bound & 0.3 \\
batch size & 16 utterances \\
number of local epochs & 1 \\
NCE noise size & 1024 \\
\bottomrule
\end{tabular}
\caption{Common hyper-parameters shared through out the experiments}
\label{tab:hyp}
\end{table}

\section{Experiments}

\subsection{Stack Overflow}
\label{subsec:stackoverflow}

\begin{table}[t]
\centering
\begin{tabular}{l|ccc}
\toprule
& train & dev & test \\
\midrule
distinct users &  342k & 38k & 204k \\
examples & 135m & 16m & 16m \\ 
\bottomrule
\end{tabular}
\caption{Stack Overflow dataset summary}
\label{tab:sof-stat}
\end{table}

\begin{table}
\centering
\begin{tabular}{m{0.2\textwidth}|p{0.07\textwidth}p{0.07\textwidth}p{0.05\textwidth}}
\toprule
\textbf{setup} & \textbf{payload} & \textbf{PPL} \\
\midrule
baseline & - & 50.96 \\
\midrule
full model & 112M & 66.47 \\
LoRA(48) & 21M & 85.24 \\
LoRA(64) & 28M & 80.13 \\
PEU(5k) & \textbf{16M} & \textbf{64.77} \\
PEU(10k) & 21M & 65.06 \\
PEU(20k) & 31M & 65.97 \\
PEU(5k)+LoRA(64) & 3.6M & 77.79 \\
PEU(10k)+LoRA(64) & 4.8M & 77.43 \\
\bottomrule
\end{tabular}
\caption{Performance on Stack Overflow.}
\label{tab:stackoverflow}
\end{table}

This dataset is derived from Stack Overflow \cite{tff/stackoverflow}.
The dataset is preprocessed by Tensorflow API \cite{Abadi_TensorFlow_Large-scale_machine_2015}.
Its statistics are summarized in Table \ref{tab:sof-stat}.
We build a 3rd-order FOFE \cite{DBLP:conf/acl/ZhangJXHD15} baseline from scratch with BMSGD \cite{DBLP:conf/icassp/ChenH16}. We apply our proposed methods in the PFL context.
We summarize the hyper-parameters in Table \ref{tab:hyp}.
The model includes an IO-tied \cite{DBLP:conf/eacl/PressW17} 256-dimension word embedding of 100k vocabulary and 4 fully-connected layers of 768 nodes.


\subsection{Speech Recognition on Virtual Assistant}
\label{subsec:msg}

\begin{table}
\centering
\begin{tabular}{m{0.3\textwidth}|p{0.044\textwidth}p{0.042\textwidth}}
\toprule
\textbf{setup} & \textbf{PPL}  & \textbf{WER} \\
\midrule
baseline & 67.33 & 3.97 \\
am-ref & 71.81 & 4.29 \\
\midrule
full model& 67.19 & 4.09 \\
LoRA(48) & 67.67 & 4.10 \\
LoRA(64) & 67.56 & 4.10 \\
PEU(5K) & 67.12 & 4.08 \\
PEU(10K) & 67.10 & 4.09 \\
PEU(20K) & 67.04 & 4.08 \\
PEU(10k) + LoRA(64) & 67.47 & 4.09 \\
\midrule
\textbf{\small real device} PEU(10k) & 67.60 & 4.13 \\
\textbf{\small real device} PEU{(10k)}+LoRA{(64)} & 67.46 & 4.12 \\
\bottomrule
\end{tabular}
\caption{Performance on randomly sampled and anonymized virtual assistant queries. Payload is same as Table \ref{tab:stackoverflow}.}
\label{tab:msg}
\end{table}


In addition to next word prediction, we evaluate our approach on automatic speech recognition (ASR), which is measured by word error rate (WER).
We experiment with a hybrid ASR system that consists of acoustic model (AM), pronunciation dictionary and LM.
Each component is represented as a WFST \cite{DBLP:journals/csl/MohriPR02} and the recognition process is formulated as a beam search on the composed WFSTs.
The training corpus contains a large number of randomly sampled and anonymized automatically-transcribed virtual assistant queries and is supplemented with templated utterances based on knowledge base statistics. 
We sampled 750 million sentences.
We curated 30 million anonymized human-labeled virtual assistant queries, from which we sampled 20k queries to form a PPL test set and another 20k to form a WER test set.
The remaining are used as AM training data.

The baseline is trained identically to Section \ref{subsec:stackoverflow} except that virtual assistant data is used.
We build a weaker baseline that uses only the AM training data, referred to as \textit{am-ref}.
This will also serve as the initial checkpoint for PFL experiments.
We follow \citet{DBLP:conf/icassp/0001NLMZL20}'s recipe to produce an AM, which is used in all experiments.
Specifically, no ngram-LM is interpolated with NNLM.
We start with simulations. 
To mimic the non-IID nature of the data present in FL, we separate the data into 500k partitions with the algorithm described in \cite{DBLP:journals/corr/abs-1909-06335} and treat each partition as an individual user device.
Finally, we replicate the results of the simulation on real user devices.


\subsection{Result Analysis}

Results of various configurations are summarized in Table \ref{tab:stackoverflow} and \ref{tab:msg}. 
Additional results for EMA comparison are described in Appendix~\ref{sec:appendix}.
For both datasets, \textit{PEU}-enabled settings converge to similar or better performance as their disabled counterparts.
Particularly in Stack Overflow, it outperforms \textit{LoRA} significantly with the same payload budget, which shows that \textit{PEU} is more suitable for training from scratch. 
Compared to full model update, \textit{PEU} dramatically reduces the memory footprint and payload.
More importantly, because the reduction is irrelevant to the prediction space, larger models potentially benefit more from \textit{PEU}.
\textit{PEU} is compatible with \textit{LoRA}.
In virtual assistant queries, the combined setting drops the payload to less than 5\% of the full model update, while maintaining accuracy of the non-private server-based baseline.
Finally, we replicate the results of the simulation on real user devices.
The results show that our proposed method is not only theoretically sound, but also feasible in production.

\section{Conclusion}

With partial embedding updates, low-rank adaptation, noise contrastive estimation and exponential moving average on trained parameters, we train large-vocabulary language models using private federated learning on resource-constrained mobile devices and achieve accuracy parity.
Our approaches are attractive for applications that need to learn a language model from user devices in a privacy-preserving manner.

\bibliography{emnlp2022}

\begin{thebibliography}{32}
\expandafter\ifx\csname natexlab\endcsname\relax\def\natexlab#1{#1}\fi

\bibitem[{Abadi et~al.(2016)Abadi, Chu, Goodfellow, McMahan, Mironov, Talwar,
  and Zhang}]{DBLP:conf/ccs/AbadiCGMMT016}
Mart{\'{\i}}n Abadi, Andy Chu, Ian~J. Goodfellow, H.~Brendan McMahan, Ilya
  Mironov, Kunal Talwar, and Li~Zhang. 2016.
\newblock \href {https://doi.org/10.1145/2976749.2978318} {Deep learning with
  differential privacy}.
\newblock In \emph{Proceedings of the 2016 {ACM} {SIGSAC} Conference on
  Computer and Communications Security, Vienna, Austria, October 24-28, 2016},
  pages 308--318. {ACM}.

\bibitem[{Abadi et~al.(2015)Abadi, Agarwal, Barham, Brevdo, Chen, Citro,
  Corrado, Davis, Dean, Devin, Ghemawat, Goodfellow, Harp, Irving, Isard,
  Jozefowicz, Jia, Kaiser, Kudlur, Levenberg, Mané, Schuster, Monga, Moore,
  Murray, Olah, Shlens, Steiner, Sutskever, Talwar, Tucker, Vanhoucke,
  Vasudevan, Viégas, Vinyals, Warden, Wattenberg, Wicke, Yu, and
  Zheng}]{Abadi_TensorFlow_Large-scale_machine_2015}
Martín Abadi, Ashish Agarwal, Paul Barham, Eugene Brevdo, Zhifeng Chen, Craig
  Citro, Greg~S. Corrado, Andy Davis, Jeffrey Dean, Matthieu Devin, Sanjay
  Ghemawat, Ian Goodfellow, Andrew Harp, Geoffrey Irving, Michael Isard, Rafal
  Jozefowicz, Yangqing Jia, Lukasz Kaiser, Manjunath Kudlur, Josh Levenberg,
  Dan Mané, Mike Schuster, Rajat Monga, Sherry Moore, Derek Murray, Chris
  Olah, Jonathon Shlens, Benoit Steiner, Ilya Sutskever, Kunal Talwar, Paul
  Tucker, Vincent Vanhoucke, Vijay Vasudevan, Fernanda Viégas, Oriol Vinyals,
  Pete Warden, Martin Wattenberg, Martin Wicke, Yuan Yu, and Xiaoqiang Zheng.
  2015.
\newblock \href {https://doi.org/10.5281/zenodo.4724125} {{TensorFlow,
  Large-scale machine learning on heterogeneous systems}}.

\bibitem[{Authors(2019)}]{tff/stackoverflow}
The Tensorflow~Federated Authors. 2019.
\newblock \href
  {https://www.tensorflow.org/federated/api_docs/python/tff/simulation/datasets/stackoverflow/load_data}
  {tff.simulation.datasets.stackoverflow.load\_data}.

\bibitem[{Bassily et~al.(2014)Bassily, Smith, and
  Thakurta}]{bassily2014private}
Raef Bassily, Adam Smith, and Abhradeep Thakurta. 2014.
\newblock Private empirical risk minimization: Efficient algorithms and tight
  error bounds.
\newblock In \emph{2014 IEEE 55th annual symposium on foundations of computer
  science}, pages 464--473. IEEE.

\bibitem[{Brown et~al.(2020)Brown, Mann, Ryder, Subbiah, Kaplan, Dhariwal,
  Neelakantan, Shyam, Sastry, Askell, Agarwal, Herbert{-}Voss, Krueger,
  Henighan, Child, Ramesh, Ziegler, Wu, Winter, Hesse, Chen, Sigler, Litwin,
  Gray, Chess, Clark, Berner, McCandlish, Radford, Sutskever, and
  Amodei}]{DBLP:conf/nips/BrownMRSKDNSSAA20}
Tom~B. Brown, Benjamin Mann, Nick Ryder, Melanie Subbiah, Jared Kaplan,
  Prafulla Dhariwal, Arvind Neelakantan, Pranav Shyam, Girish Sastry, Amanda
  Askell, Sandhini Agarwal, Ariel Herbert{-}Voss, Gretchen Krueger, Tom
  Henighan, Rewon Child, Aditya Ramesh, Daniel~M. Ziegler, Jeffrey Wu, Clemens
  Winter, Christopher Hesse, Mark Chen, Eric Sigler, Mateusz Litwin, Scott
  Gray, Benjamin Chess, Jack Clark, Christopher Berner, Sam McCandlish, Alec
  Radford, Ilya Sutskever, and Dario Amodei. 2020.
\newblock \href
  {https://proceedings.neurips.cc/paper/2020/hash/1457c0d6bfcb4967418bfb8ac142f64a-Abstract.html}
  {Language models are few-shot learners}.
\newblock In \emph{Advances in Neural Information Processing Systems 33: Annual
  Conference on Neural Information Processing Systems 2020, NeurIPS 2020,
  December 6-12, 2020, virtual}.

\bibitem[{Carlini et~al.(2019)Carlini, Liu, Erlingsson, Kos, and
  Song}]{DBLP:conf/uss/Carlini0EKS19}
Nicholas Carlini, Chang Liu, {\'{U}}lfar Erlingsson, Jernej Kos, and Dawn Song.
  2019.
\newblock \href
  {https://www.usenix.org/conference/usenixsecurity19/presentation/carlini}
  {The secret sharer: Evaluating and testing unintended memorization in neural
  networks}.
\newblock In \emph{28th {USENIX} Security Symposium, {USENIX} Security 2019,
  Santa Clara, CA, USA, August 14-16, 2019}, pages 267--284. {USENIX}
  Association.

\bibitem[{Carlini et~al.(2021)Carlini, Tram{\`{e}}r, Wallace, Jagielski,
  Herbert{-}Voss, Lee, Roberts, Brown, Song, Erlingsson, Oprea, and
  Raffel}]{DBLP:conf/uss/CarliniTWJHLRBS21}
Nicholas Carlini, Florian Tram{\`{e}}r, Eric Wallace, Matthew Jagielski, Ariel
  Herbert{-}Voss, Katherine Lee, Adam Roberts, Tom~B. Brown, Dawn Song,
  {\'{U}}lfar Erlingsson, Alina Oprea, and Colin Raffel. 2021.
\newblock \href
  {https://www.usenix.org/conference/usenixsecurity21/presentation/carlini-extracting}
  {Extracting training data from large language models}.
\newblock In \emph{30th {USENIX} Security Symposium, {USENIX} Security 2021,
  August 11-13, 2021}, pages 2633--2650. {USENIX} Association.

\bibitem[{Chen and Huo(2016)}]{DBLP:conf/icassp/ChenH16}
Kai Chen and Qiang Huo. 2016.
\newblock \href {https://doi.org/10.1109/ICASSP.2016.7472805} {Scalable
  training of deep learning machines by incremental block training with
  intra-block parallel optimization and blockwise model-update filtering}.
\newblock In \emph{2016 {IEEE} International Conference on Acoustics, Speech
  and Signal Processing, {ICASSP} 2016, Shanghai, China, March 20-25, 2016},
  pages 5880--5884. {IEEE}.

\bibitem[{Devlin et~al.(2019)Devlin, Chang, Lee, and
  Toutanova}]{DBLP:conf/naacl/DevlinCLT19}
Jacob Devlin, Ming{-}Wei Chang, Kenton Lee, and Kristina Toutanova. 2019.
\newblock \href {https://doi.org/10.18653/v1/n19-1423} {{BERT:} pre-training of
  deep bidirectional transformers for language understanding}.
\newblock In \emph{Proceedings of the 2019 Conference of the North American
  Chapter of the Association for Computational Linguistics: Human Language
  Technologies, {NAACL-HLT} 2019, Minneapolis, MN, USA, June 2-7, 2019, Volume
  1 (Long and Short Papers)}, pages 4171--4186. Association for Computational
  Linguistics.

\bibitem[{Dwork et~al.(2006)Dwork, McSherry, Nissim, and
  Smith}]{DBLP:conf/tcc/DworkMNS06}
Cynthia Dwork, Frank McSherry, Kobbi Nissim, and Adam~D. Smith. 2006.
\newblock \href {https://doi.org/10.1007/11681878\_14} {Calibrating noise to
  sensitivity in private data analysis}.
\newblock In \emph{Theory of Cryptography, Third Theory of Cryptography
  Conference, {TCC} 2006, New York, NY, USA, March 4-7, 2006, Proceedings},
  volume 3876 of \emph{Lecture Notes in Computer Science}, pages 265--284.
  Springer.

\bibitem[{Dwork and Roth(2014)}]{DBLP:journals/fttcs/DworkR14}
Cynthia Dwork and Aaron Roth. 2014.
\newblock \href {https://doi.org/10.1561/0400000042} {The algorithmic
  foundations of differential privacy}.
\newblock \emph{Found. Trends Theor. Comput. Sci.}, 9(3-4):211--407.

\bibitem[{Gutmann and Hyv{\"{a}}rinen(2010)}]{DBLP:journals/jmlr/GutmannH10}
Michael Gutmann and Aapo Hyv{\"{a}}rinen. 2010.
\newblock \href {http://proceedings.mlr.press/v9/gutmann10a.html}
  {Noise-contrastive estimation: {A} new estimation principle for unnormalized
  statistical models}.
\newblock In \emph{Proceedings of the Thirteenth International Conference on
  Artificial Intelligence and Statistics, {AISTATS} 2010, Chia Laguna Resort,
  Sardinia, Italy, May 13-15, 2010}, volume~9 of \emph{{JMLR} Proceedings},
  pages 297--304. JMLR.org.

\bibitem[{Hsu et~al.(2019)Hsu, Qi, and
  Brown}]{DBLP:journals/corr/abs-1909-06335}
Tzu{-}Ming~Harry Hsu, Hang Qi, and Matthew Brown. 2019.
\newblock \href {http://arxiv.org/abs/1909.06335} {Measuring the effects of
  non-identical data distribution for federated visual classification}.
\newblock \emph{CoRR}, abs/1909.06335.

\bibitem[{Hu et~al.(2021)Hu, Shen, Wallis, Allen{-}Zhu, Li, Wang, and
  Chen}]{DBLP:journals/corr/abs-2106-09685}
Edward~J. Hu, Yelong Shen, Phillip Wallis, Zeyuan Allen{-}Zhu, Yuanzhi Li,
  Shean Wang, and Weizhu Chen. 2021.
\newblock \href {http://arxiv.org/abs/2106.09685} {Lora: Low-rank adaptation of
  large language models}.
\newblock \emph{CoRR}, abs/2106.09685.

\bibitem[{Huang et~al.(2020)Huang, Ng, Liu, Mason, Zhuang, and
  Liu}]{DBLP:conf/icassp/0001NLMZL20}
Zhen Huang, Tim Ng, Leo Liu, Henry Mason, Xiaodan Zhuang, and Daben Liu. 2020.
\newblock \href {https://doi.org/10.1109/ICASSP40776.2020.9053973} {{SNDCNN:}
  self-normalizing deep cnns with scaled exponential linear units for speech
  recognition}.
\newblock In \emph{2020 {IEEE} International Conference on Acoustics, Speech
  and Signal Processing, {ICASSP} 2020, Barcelona, Spain, May 4-8, 2020}, pages
  6854--6858. {IEEE}.

\bibitem[{Martens(2020)}]{DBLP:journals/jmlr/Martens20}
James Martens. 2020.
\newblock \href {http://jmlr.org/papers/v21/17-678.html} {New insights and
  perspectives on the natural gradient method}.
\newblock \emph{J. Mach. Learn. Res.}, 21:146:1--146:76.

\bibitem[{McMahan et~al.(2017)McMahan, Moore, Ramage, Hampson, and
  y~Arcas}]{DBLP:conf/aistats/McMahanMRHA17}
Brendan McMahan, Eider Moore, Daniel Ramage, Seth Hampson, and
  Blaise~Ag{\"{u}}era y~Arcas. 2017.
\newblock \href {http://proceedings.mlr.press/v54/mcmahan17a.html}
  {Communication-efficient learning of deep networks from decentralized data}.
\newblock In \emph{Proceedings of the 20th International Conference on
  Artificial Intelligence and Statistics, {AISTATS} 2017, 20-22 April 2017,
  Fort Lauderdale, FL, {USA}}, volume~54 of \emph{Proceedings of Machine
  Learning Research}, pages 1273--1282. {PMLR}.

\bibitem[{McMahan et~al.(2018)McMahan, Ramage, Talwar, and
  Zhang}]{DBLP:conf/iclr/McMahanRT018}
H.~Brendan McMahan, Daniel Ramage, Kunal Talwar, and Li~Zhang. 2018.
\newblock \href {https://openreview.net/forum?id=BJ0hF1Z0b} {Learning
  differentially private recurrent language models}.
\newblock In \emph{6th International Conference on Learning Representations,
  {ICLR} 2018, Vancouver, BC, Canada, April 30 - May 3, 2018, Conference Track
  Proceedings}. OpenReview.net.

\bibitem[{Melis et~al.(2019)Melis, Song, Cristofaro, and
  Shmatikov}]{DBLP:conf/sp/MelisSCS19}
Luca Melis, Congzheng Song, Emiliano~De Cristofaro, and Vitaly Shmatikov. 2019.
\newblock \href {https://doi.org/10.1109/SP.2019.00029} {Exploiting unintended
  feature leakage in collaborative learning}.
\newblock In \emph{2019 {IEEE} Symposium on Security and Privacy, {SP} 2019,
  San Francisco, CA, USA, May 19-23, 2019}, pages 691--706. {IEEE}.

\bibitem[{Mironov(2017)}]{DBLP:conf/csfw/Mironov17}
Ilya Mironov. 2017.
\newblock \href {https://doi.org/10.1109/CSF.2017.11} {R{\'{e}}nyi differential
  privacy}.
\newblock In \emph{30th {IEEE} Computer Security Foundations Symposium, {CSF}
  2017, Santa Barbara, CA, USA, August 21-25, 2017}, pages 263--275. {IEEE}
  Computer Society.

\bibitem[{Mironov et~al.(2019)Mironov, Talwar, and
  Zhang}]{DBLP:journals/corr/abs-1908-10530}
Ilya Mironov, Kunal Talwar, and Li~Zhang. 2019.
\newblock \href {http://arxiv.org/abs/1908.10530} {R{\'{e}}nyi differential
  privacy of the sampled gaussian mechanism}.
\newblock \emph{CoRR}, abs/1908.10530.

\bibitem[{Mnih and Teh(2012)}]{DBLP:conf/icml/MnihT12}
Andriy Mnih and Yee~Whye Teh. 2012.
\newblock \href {http://icml.cc/2012/papers/855.pdf} {A fast and simple
  algorithm for training neural probabilistic language models}.
\newblock In \emph{Proceedings of the 29th International Conference on Machine
  Learning, {ICML} 2012, Edinburgh, Scotland, UK, June 26 - July 1, 2012}.
  icml.cc / Omnipress.

\bibitem[{Mohri et~al.(2002)Mohri, Pereira, and
  Riley}]{DBLP:journals/csl/MohriPR02}
Mehryar Mohri, Fernando Pereira, and Michael Riley. 2002.
\newblock \href {https://doi.org/10.1006/csla.2001.0184} {Weighted finite-state
  transducers in speech recognition}.
\newblock \emph{Comput. Speech Lang.}, 16(1):69--88.

\bibitem[{Press and Wolf(2017)}]{DBLP:conf/eacl/PressW17}
Ofir Press and Lior Wolf. 2017.
\newblock \href {https://doi.org/10.18653/v1/e17-2025} {Using the output
  embedding to improve language models}.
\newblock In \emph{Proceedings of the 15th Conference of the European Chapter
  of the Association for Computational Linguistics, {EACL} 2017, Valencia,
  Spain, April 3-7, 2017, Volume 2: Short Papers}, pages 157--163. Association
  for Computational Linguistics.

\bibitem[{Radford et~al.(2018)Radford, Narasimhan, Salimans, and
  Sutskever}]{radford2018improving}
Alec Radford, Karthik Narasimhan, Tim Salimans, and Ilya Sutskever. 2018.
\newblock Improving language understanding by generative pre-training.

\bibitem[{Radford et~al.(2019)Radford, Wu, Child, Luan, Amodei, Sutskever
  et~al.}]{radford2019language}
Alec Radford, Jeffrey Wu, Rewon Child, David Luan, Dario Amodei, Ilya
  Sutskever, et~al. 2019.
\newblock Language models are unsupervised multitask learners.
\newblock \emph{OpenAI blog}, 1(8):9.

\bibitem[{Reddi et~al.(2021)Reddi, Charles, Zaheer, Garrett, Rush,
  Kone{\v{c}}n{\'y}, Kumar, and McMahan}]{DBLP:conf/iclr/ReddiCZGRKKM21}
Sashank~J. Reddi, Zachary Charles, Manzil Zaheer, Zachary Garrett, Keith Rush,
  Jakub Kone{\v{c}}n{\'y}, Sanjiv Kumar, and Hugh~Brendan McMahan. 2021.
\newblock \href {https://openreview.net/forum?id=LkFG3lB13U5} {Adaptive
  federated optimization}.
\newblock In \emph{9th International Conference on Learning Representations,
  {ICLR} 2021, Virtual Event, Austria, May 3-7, 2021}. OpenReview.net.

\bibitem[{Saebi et~al.(2021)Saebi, Pusateri, Meghawat, and
  Gysel}]{DBLP:conf/interspeech/SaebiPMG21}
Mandana Saebi, Ernest Pusateri, Aaksha Meghawat, and Christophe~Van Gysel.
  2021.
\newblock \href {https://doi.org/10.21437/Interspeech.2021-1767} {A
  discriminative entity-aware language model for virtual assistants}.
\newblock In \emph{Interspeech 2021, 22nd Annual Conference of the
  International Speech Communication Association, Brno, Czechia, 30 August - 3
  September 2021}, pages 2032--2036. {ISCA}.

\bibitem[{Shoeybi et~al.(2019)Shoeybi, Patwary, Puri, LeGresley, Casper, and
  Catanzaro}]{DBLP:journals/corr/abs-1909-08053}
Mohammad Shoeybi, Mostofa Patwary, Raul Puri, Patrick LeGresley, Jared Casper,
  and Bryan Catanzaro. 2019.
\newblock \href {http://arxiv.org/abs/1909.08053} {Megatron-lm: Training
  multi-billion parameter language models using model parallelism}.
\newblock \emph{CoRR}, abs/1909.08053.

\bibitem[{Vaswani et~al.(2017)Vaswani, Shazeer, Parmar, Uszkoreit, Jones,
  Gomez, Kaiser, and Polosukhin}]{DBLP:conf/nips/VaswaniSPUJGKP17}
Ashish Vaswani, Noam Shazeer, Niki Parmar, Jakob Uszkoreit, Llion Jones,
  Aidan~N. Gomez, Lukasz Kaiser, and Illia Polosukhin. 2017.
\newblock \href
  {https://proceedings.neurips.cc/paper/2017/hash/3f5ee243547dee91fbd053c1c4a845aa-Abstract.html}
  {Attention is all you need}.
\newblock In \emph{Advances in Neural Information Processing Systems 30: Annual
  Conference on Neural Information Processing Systems 2017, December 4-9, 2017,
  Long Beach, CA, {USA}}, pages 5998--6008.

\bibitem[{Zhang et~al.(2015)Zhang, Jiang, Xu, Hou, and
  Dai}]{DBLP:conf/acl/ZhangJXHD15}
Shiliang Zhang, Hui Jiang, Mingbin Xu, Junfeng Hou, and Li{-}Rong Dai. 2015.
\newblock \href {https://doi.org/10.3115/v1/p15-2081} {The fixed-size
  ordinally-forgetting encoding method for neural network language models}.
\newblock In \emph{Proceedings of the 53rd Annual Meeting of the Association
  for Computational Linguistics and the 7th International Joint Conference on
  Natural Language Processing of the Asian Federation of Natural Language
  Processing, {ACL} 2015, July 26-31, 2015, Beijing, China, Volume 2: Short
  Papers}, pages 495--500. The Association for Computer Linguistics.

\bibitem[{Zhu et~al.(2019)Zhu, Liu, and Han}]{DBLP:conf/nips/ZhuLH19}
Ligeng Zhu, Zhijian Liu, and Song Han. 2019.
\newblock \href
  {https://proceedings.neurips.cc/paper/2019/hash/60a6c4002cc7b29142def8871531281a-Abstract.html}
  {Deep leakage from gradients}.
\newblock In \emph{Advances in Neural Information Processing Systems 32: Annual
  Conference on Neural Information Processing Systems 2019, NeurIPS 2019,
  December 8-14, 2019, Vancouver, BC, Canada}, pages 14747--14756.

\end{thebibliography}
\bibliographystyle{acl_natbib}

\newpage
\appendix
\begin{figure}
    \centering
    \includegraphics[width=\linewidth]{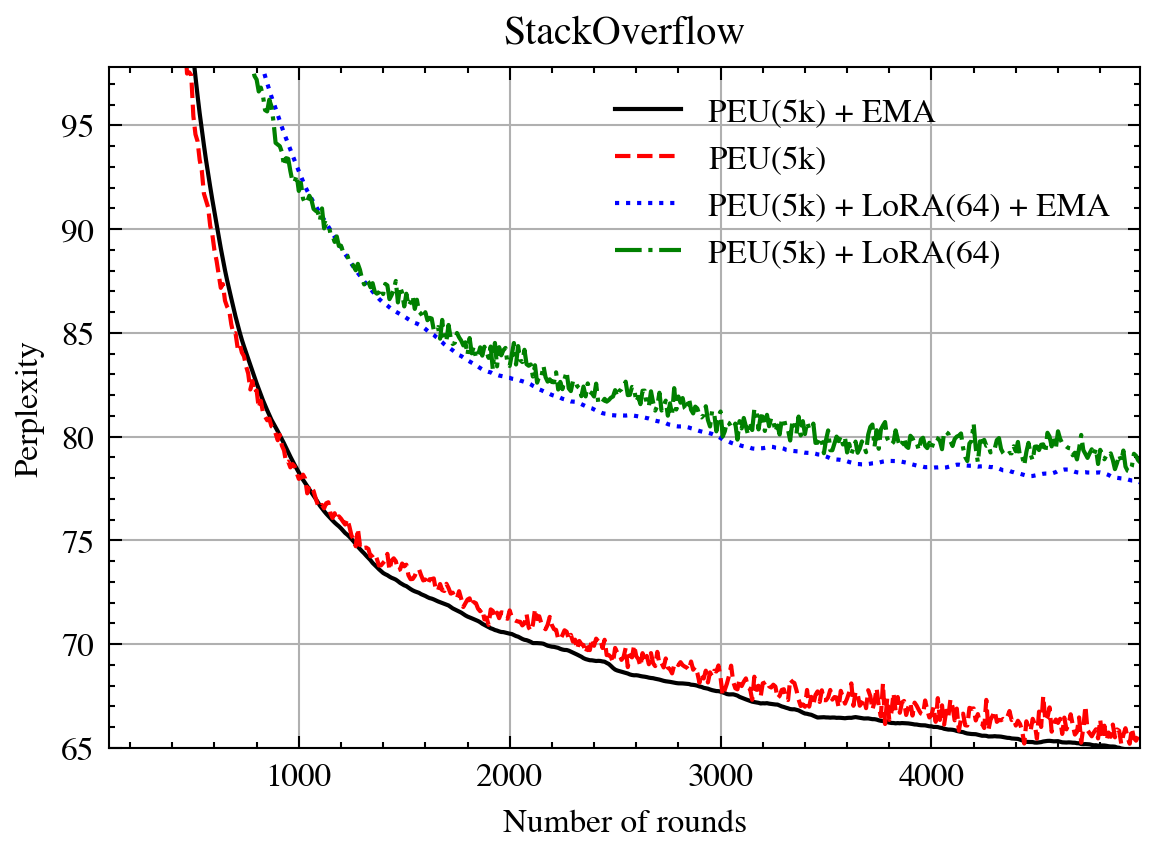}
    \includegraphics[width=\linewidth]{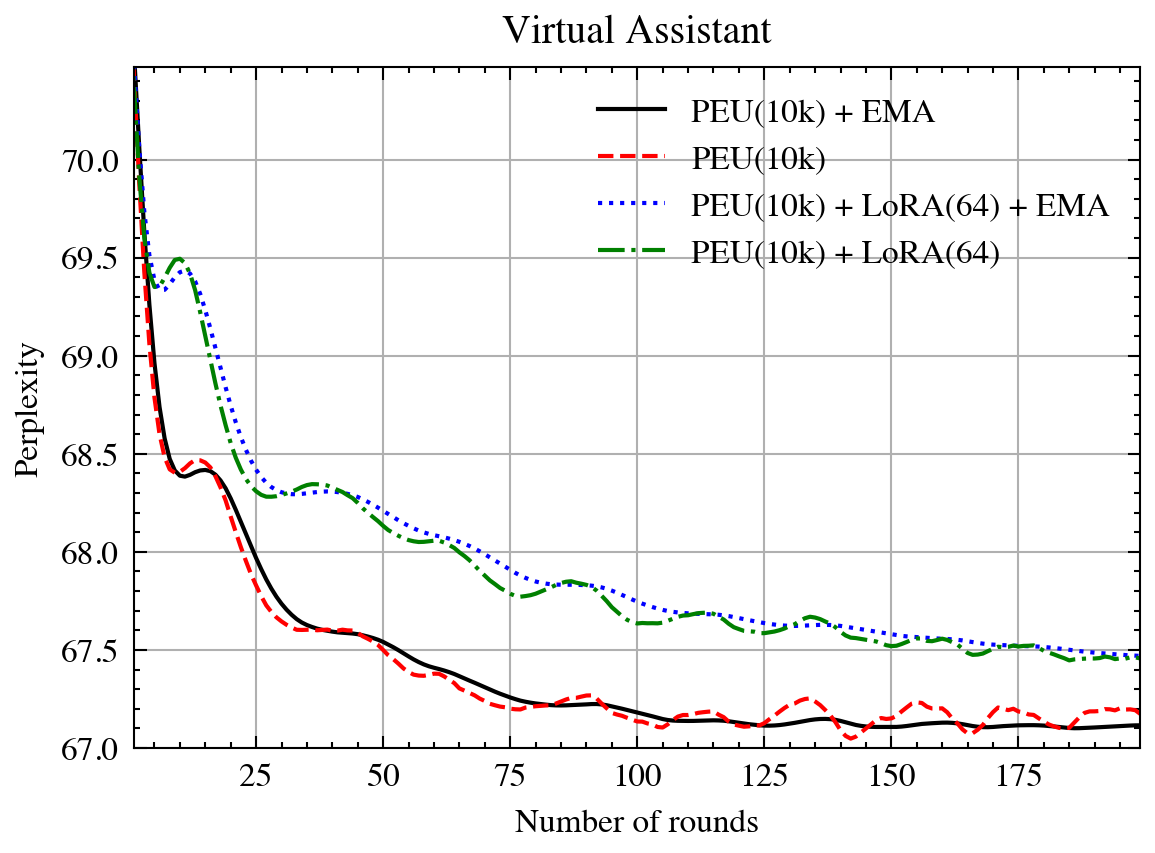}
    \caption{Effect of EMA ($\gamma=0.999$) with \emph{PEU} and \emph{PEU + LoRA} setup. x-axis is the number of PFL rounds and y-axis is the perplexity on the dev set. }
    \label{fig:ema}
\end{figure}
\section{Additional Results}
\label{sec:appendix}

Figure~\ref{fig:ema} demonstrates the effect of EMA described in Section~\ref{sec:ema}. 
For both \emph{PEU} and \emph{PEU + LoRA} setup on the StackOverflow dataset, EMA models achieved better perplexity and their convergence curves are much smoother due to the reason that moving average can smooth the DP noise. 
For virtual assistant task, EMA models did produce a stabler curve but the perplexity values are similar to models without EMA.
We suspect it is because that virtual assistant models started from a pretrained model which was already close to a local minima and there is a little room for EMA to improve the convergence in this case.
\end{document}